\documentclass[journal,transmag]{IEEEtran}
\usepackage{cite}
\usepackage{xcolor}

\ifCLASSINFOpdf
   \usepackage[pdftex]{graphicx}
\else
   \usepackage[dvips]{graphicx}
\fi
\usepackage{amsmath}
\usepackage{amssymb}
\usepackage{multirow}
\newcommand{\norm}[1]{\left\lVert#1\right\rVert}
\ifCLASSOPTIONcompsoc
\usepackage[caption=false,font=normalsize,labelfont=sf,textfont=sf]{subfig}
\else
  \usepackage[caption=false,font=footnotesize]{subfig}
\fi
\begin{document}
%
\title{Be Your Own Best Competitor! \\  Multi-Branched Adversarial Knowledge Transfer}


\author{\IEEEauthorblockN{Mahdi Ghorbani\IEEEauthorrefmark{1},
Fahimeh Fooladgar\IEEEauthorrefmark{1}, and
Shohreh Kasaei\IEEEauthorrefmark{2},~\IEEEmembership{Senior Member,~IEEE}}
\IEEEauthorblockA{Department of Computer Engineering,
Sharif University of Technology, Tehran, Iran}
\thanks{\IEEEauthorrefmark{1}The first two authors contributed equally to this work. 

\IEEEauthorrefmark{2}Corresponding author (email: kasaei@sharif.edu).}}

%



\IEEEtitleabstractindextext{%
\begin{abstract}
Deep neural network architectures have attained remarkable improvements in scene understanding tasks (such as image classification and semantic segmentation). Utilizing an efficient model is one of the most important constraints for limited-resource devices. Recently, several compression methods have been proposed to diminish the heavy computational burden and memory consumption. Among them, the pruning and quantizing methods exhibit a critical drop in performances by compressing the model parameters. While the knowledge distillation methods improve the performance of compact models by focusing on training lightweight networks with the supervision of cumbersome networks.  In the proposed method, the knowledge distillation has been performed within the network by constructing multiple branches over the primary stream of the model, known as the self-distillation method.  Therefore, the ensemble of sub-neural network models has been proposed to transfer the knowledge among themselves with the knowledge distillation policies as well as an adversarial learning strategy. Hence, The proposed ensemble of sub-models is trained against a discriminator model adversarially. Besides, their knowledge is transferred within the ensemble by four different loss functions. The proposed method has been devoted to both lightweight image classification and encoder-decoder architectures to boost the performance of small and compact models without incurring extra computational overhead at the inference process. Extensive experimental results on the main challenging datasets show that the proposed network outperforms the primary model in terms of accuracy at the same number of parameters and computational cost. The obtained results show that the proposed model has achieved significant improvement over earlier ideas of self-distillation methods. The effectiveness of the proposed models has also been illustrated in the encoder-decoder model.
\end{abstract}

}

\maketitle

\IEEEdisplaynontitleabstractindextext

%
\IEEEpeerreviewmaketitle

\section{Introduction}
\IEEEPARstart{D}{EEP}  Convolutional Neural Networks (CNNs) have been adopted in most categories of visual recognition tasks (such as image classification \cite{he2016deep,he2016identity,hu2018squeeze} and semantic segmentation \cite{naseer2018indoor,long2015fully,fooladgar2019survey}). The state-of-the-art models in terms of accuracy have been designed through deeper and more complex network architectures to learn a large number of parameters via a high number of computational operations. In some applications,  with limited resource devices (such as autonomous driving \cite{grigorescu2019survey,kuutti2020survey} and connected vehicle control\cite{alam2019taawun}), not only the high level of model accuracy is a substantial factor, but also the model size, inference time, and computational cost are significantly important. 

To design more efficient models in terms of these costs, some researchers have investigated the amount of redundancy in the weights of deep neural networks. Therefore, novel factorization \cite{denton2014exploiting}, pruning \cite{han2015learning,li2016pruning}, quantization \cite{chen2015compressing,courbariaux2016binarized}, and compression methods \cite{hinton2015distilling,ba2014deep} have been proposed. These pruning/quantizing methods decreased the run-time memory and computational costs at a moderate accuracy loss. Among them, the Knowledge Distillation  (KD) methods are considered as a compression strategy with a different and novel insight. 

The main idea of knowledge distillation is to transfer the generalization ability of the network from a large and more complex network (known as teacher) to a lightweight and smaller network (known as student). This transfer strategy can lead to boost the accuracy of smaller models. Interestingly, the student might even infrequently attain better accuracy than the teacher. Consequently, the cumbersome network has been superseded by the compact student network with lower computational cost, shorter inference time, and smaller model size properties. This transferred information between teacher and student can be performed on the softmax layer of the network where the student tries to estimate the softer output of the final layer of its teacher instead of the hard target of ground-truth \cite{hinton2015distilling}. Also, the information between each layer of the teacher network can be transferred to the student model via minimizing the distance among feature maps of teacher and student model in intermediate layers \cite{romero2014fitnets}.

It is important how the knowledge is extracted from the teacher model and how it can be transferred to the student model. One of the main challenges of knowledge distillation is how teacher-student architecture should be designed.  Finding a proper teacher network for each student model where it can learn significantly is still a challenging process.  The capacity gap between the teacher and student model declines the transferred knowledge; hence the student cannot be effectively taught by the teacher model \cite{zhang2019your}. 
Zhang et al. \cite{zhang2019your} proposed one-step knowledge distillation where the knowledge was transferred among different levels of the model to squeeze the information of deeper layers into early layers. This method, known as self-distillation, increased the generalization ability of the model; hence it achieved better accuracy on the test data.

On the other hand, some especial model structures with outstanding architecture designs have been proposed to deal with computational complexity and model size.  MobileNet \cite{howard2017mobilenets}, ShuffleNet \cite{zhang2018shufflenet}, and CondenseNet \cite{huang2018condensenet} reduced the computational cost while obtaining a comparable accuracy to the VGG \cite{simonyan2014very} on the ImageNet \cite{krizhevsky2012imagenet} dataset. 

 In this paper,  the knowledge distillation has been performed within the network from the deeper level into the shallower level by ensemble and adversarial learning ideas. Generative Adversarial Networks (GANs) are consist of two different deep neural networks (called generator and discriminator) competing with each other. The goal of the generator is to mislead the discriminator by generating outputs that look real, while the discriminator tries to distinguish between the real and generated fake outputs. Both networks get trained by going back and forth through this procedure. 

In the proposed method, multiple generators have been trained adversarially against a single discriminator.  They are branched over the original stream of a primary network. Each branch is considered as a member of the proposed ensemble of sub-neural network models (Fig. \ref{fig:class_arch}).  Based on the architecture of the primary model, the constructed branches produce different levels of sub-model (from shallower to deeper ones). The main goals are not only to improve the overall performance of these ensemble models by aggregating their predictions but also to train these sub-neural network models to learn from each other by transferring their knowledge among themselves. Hence, knowledge can be transferred in three different ways. First, it can be transferred among these sub-models by the Kulback-Leibler (KL) divergence loss between each pair of the generators' outputs. Second,  it is transferred from the deeper to the shallower sub-model by the L2 loss between similarity maps. Third, it can be transferred from the weighted average of all sub-models' output and ground-truth labels by the Wasserstein loss \cite{gulrajani2017improved}. The adversarial learning has been adopted in this framework to encourage each sub-model prediction output to be similar to the weighted average over an ensemble of generators and the real labels. Hence, the generator models are aimed to deceive a discriminator network into believing that their output labels are the real labels. 

This novel architecture has been exploited in image classification as well as semantic segmentation networks.  
To the best of our knowledge, this is the first work to propose ensemble with the adversarial learning in the knowledge distillation paradigm for semantic segmentation task. The obtained results show that the proposed distillation framework has outperformed the original primary model and has obtained higher accuracy with the same number of parameters and computational complexity at inference time.

The main contributions of this paper are as follows:
\begin{itemize}
	\item Proposing an end-to-end self-teaching framework based on adversarial learning with an ensemble of student models for both image classification and semantic segmentation tasks.
	\item Deriving benefits from ensemble learning while keeping the inference time and computational cost as same as the base model.
	\item Conducting extensive experiments on both image classification and semantic segmentation tasks to demonstrate the effectiveness of the proposed model.
\end{itemize}

This paper is organized as follows. In Section \ref{sec:2}, the related work is explained in two different categories of image classification and semantic segmentation. The proposed adversarial self-teaching model is introduced in Section \ref{sec:3}. Experimental results on well-known image labeling datasets are reported in Section \ref{sec:4}. In Section \ref{sec:5}, concluding remarks are presented.

 

\section{Related Work}
\label{sec:2}
Along with the great level of accuracy obtained via deep neural network architectures, there is some research on designing efficient models to diminish computational cost, memory requirements, and inference time of the models. The goal of designing efficient networks established several research directions. These can be classified into three main categories of (i) designing efficient architectures, (ii) learning efficient architectures, and (iii) compressing the model size.

\subsection{Image Classification}
Specific-purpose CNN models (such as CondenseNet \cite{huang2018condensenet}, MobileNet \cite{howard2017mobilenets}, ShuffleNet \cite{zhang2018shufflenet}, and RDenseCNN \cite{fooladgar2020lightweight}) have been designed to reduce the model size (number of parameters), the computational cost (number of Add-Multiplication operations), and memory requirement with the speedup in the inference time. Some of them utilize the idea of group convolution and depth-wise separable convolution \cite{howard2017mobilenets}. In \cite{huang2018condensenet}, the kernel pruning is performed as the condensation procedure by the condensation factor. 

Designing a family of models via automatically architecture search methods have been proposed to effectively scale the network's depth, width, and resolutions to achieve a higher efficiency alongside accuracy \cite{tan2019efficientnet}. BlockQNN has been proposed in \cite{zhong2020blockqnn}  by utilizing the Q-Learning paradigm with the epsilon-greedy exploration strategy to generate an efficient network, automatically. These architecture search methods impose computational power in the training phase.

In \cite{denton2014exploiting}, the fully connected weights of CNN models have been approximated by singular value decomposition methods to compress the large model. The weight quantization \cite{chen2015compressing,courbariaux2016binarized,rastegari2016xnor,jacob2018quantization} has been proposed to compress the model size and speed up the inference time. The weight, kernel, channel, and layer levels pruning methods \cite{han2015learning,li2016pruning,zhou2016less} have been presented to compress the model size and save the memory. Wen et al. \cite{wen2016learning} have applied the sparsity at different levels of CNNs' structure (kernels, channels, and layers). All of these methods have decreased the computational complexity at a moderate accuracy loss.

It is worth mentioning that a comprehensive review on the structure and learning of efficient models is outside the scope of this paper. Also, the model compression and acceleration methods in which the quantization, factorization, and pruning methods have been applied are beyond the scope of this paper. This paper focuses on the representative type of model compression known as knowledge distillation, which has concentrated overwhelming attention from the research community. Hence, they are considered and discussed as follows.

The main goal of knowledge distillation methods is to transfer knowledge of a strong learning capacity model to a lower capacity neural network structure. The early knowledge distillation method encourages the compact model to produce the output of the cumbersome model's softmax layer; instead of the hard target of ground-truth \cite{hinton2015distilling}. Ba et al. \cite{ba2014deep} demonstrated that deep neural networks might not need to be deep when it is possible to mimic the function learned by a deep and complex model with a shallower network. Consequently, they can apply model compression to train small neural networks to imitate the more complex and deeper ones.

 Yuan et al. \cite{yuan2019revisit} interpreted the knowledge distillation as a regularization method and provided the relation between knowledge distillation and label smoothing regularization, theoretically. Therefore, they proposed teacher-free knowledge distillation where the network  is trained by itself or a virtual teacher model. As such, they utilized the pre-trained model of the network as a teacher model and afterward trained this model further by itself via transferring soft targets of the pre-trained model. 


In \cite{mirzadeh2019improved}, the capacity gap between the large teacher model and the student has been investigated. It shows that the relationships between the architecture of the teacher and student model are very important. Hence, it may be a case that the student model cannot learn a significant knowledge from a specific teacher just like the habits of human beings learning where the student should find a more appropriate teacher.  

 
In the response-based knowledge distillation \cite{hinton2015distilling, ba2014deep}, the teacher model's final prediction is directly mimicked by the student model. In the feature-based knowledge distillation, the output of all intermediate layers (feature maps) can be utilized to supervise the student model \cite{romero2014fitnets}. The attention map of intermediate feature maps can be expressed as knowledge to transfer to the small student model \cite{zagoruyko2016paying}.

In \cite{gou2020survey}, the learning schemes of knowledge distillation have been classified into three categories of offline, online, and self-distillation. In the offline distillation, the knowledge is transferred from a pre-trained teacher model into a student model \cite{hinton2015distilling,romero2014fitnets,heo2019knowledge}. If the teacher model is updated simultaneously with the student model, it is considerred as online distillation \cite{chen2020online}.  In online training, multiple neural networks can work collaboratively that one of them is considered as the student model and the others are utilized as teacher models \cite{zhang2018deep}. The authors of \cite{zhu2018knowledge} proposed an online distillation in a multiple branches where each branch is considered as a student model and the overall multi-branch network is considered as their teacher. The self-distillation is a special case of online distillation where the single model are utilized as both teacher and student model. Hence, the knowledge is transferred within the model. 

The idea of multi-classifier architecture has been proposed by Deeply Supervised Net (DSN) \cite{lee2015deeply} where each classifier is trained based on the ground-truth label without employing a KD idea. In the self-distillation method of \cite{zhang2019your},  a multi-branched network has been adopted where each branch is considered as a classifier. This multi-classifier architecture is trained by utilizing the cross-entropy loss function,  the KL divergence loss between classifiers' output, and L2 loss between their feature maps. Consequently, the deepest branch guides the shallower ones by minimizing the KL and L2 losses. The Multi-self-distillation has been proposed in \cite{luan2019msd} with a similar idea of \cite{zhang2019your}, but the KL divergence has been considered among all pairs of the output of the classifiers.  

Liu et al. \cite{liu2018ktan} utilized a discriminator network as well as teacher and student generators to guide the student to simulate the output of shared knowledge extracted from the intermediate layer of the teacher model by aligning their pixel value. In \cite{shen2019meal}, the ensemble of large and complex models is distilled in a single lightweight model to incorporate the properties of ensemble learning while being computationally efficient in the test time. The authors of \cite{shen2019meal} proposed an adversarial-based learning strategy to persuade the student to generate similar outputs as ensembles of teacher models.

\subsection{Semantic Segmentation}
The popular CNN models have been mainly proposed for image classification purposes. They have been then extended to dense labeling problems such as semantic segmentation and depth estimation. Different methods have been presented to improve the segmentation outputs of these networks \cite{badrinarayanan2015segnet,noh2015learning, long2015fully,chen2016deeplab,fooladgar20193m2rnet}. All those methods incorporate the well-known CNN models in their architecture as the backbone (or encoder stream) to encode extracted features and information. By improving the performance of the backbone or utilizing a more complex and stronger encoder (like ResNet-152 and Xception \cite{chollet2017xception}), significant improvements have been achieved (such as RefineNet \cite{lin2017refinenet}, PSPNet \cite{zhao2017pyramid} and DeepLab \cite{chen2016deeplab}). But, this significant performance gain is accompanied by higher computational complexity and model parameters. 

Recently, designing efficient models in terms of model size and computational cost have concentrated on real-time and mobile applications \cite{fooladgar2019multi}. Knowledge distillation methods are also utilized in semantic segmentation to increase the accuracy of the smaller model under the guidance of stronger teacher model \cite{liu2019structured,xie2018improving}.

In comparison to designing an efficient model, the knowledge distillation methods are more promising as the semantic segmentation models are larger and more complex than the image classification task. For example, it is necessary to design both efficient encoder and decoder parts (the encoder is usually pre-trained by ImageNet). Knowledge distillation strategies can be applied in both encoder and decoder parts to improve the performance of the smaller models. The authors of \cite{liu2019structured} have proposed pixel-wise, pair-wise, and holistic distillation methods that transfer the class probabilities, similarity maps among each pair of pixel's feature vectors, and higher-order consistency of segmented outputs, respectively. In \cite{xie2018improving}, the class probability of each pixel and their local neighborhoods have been captured as the zero-order and first-order knowledge that are transferred between the  outputs of student and teacher networks. 
In \cite{shan2019distilling}, the spatial structural information has been encoded for high-level features in each pixel location in the form of pixel-wise feature similarities. Besides, the cost function has been proposed that each pixel has been selectively contributed to the loss function to avoid transferring knowledge from teacher to student model in some pixels. 

\begin{figure*}[!t]
	\centering
	\subfloat[Image classification network architecture]{
		\includegraphics[width=5in]{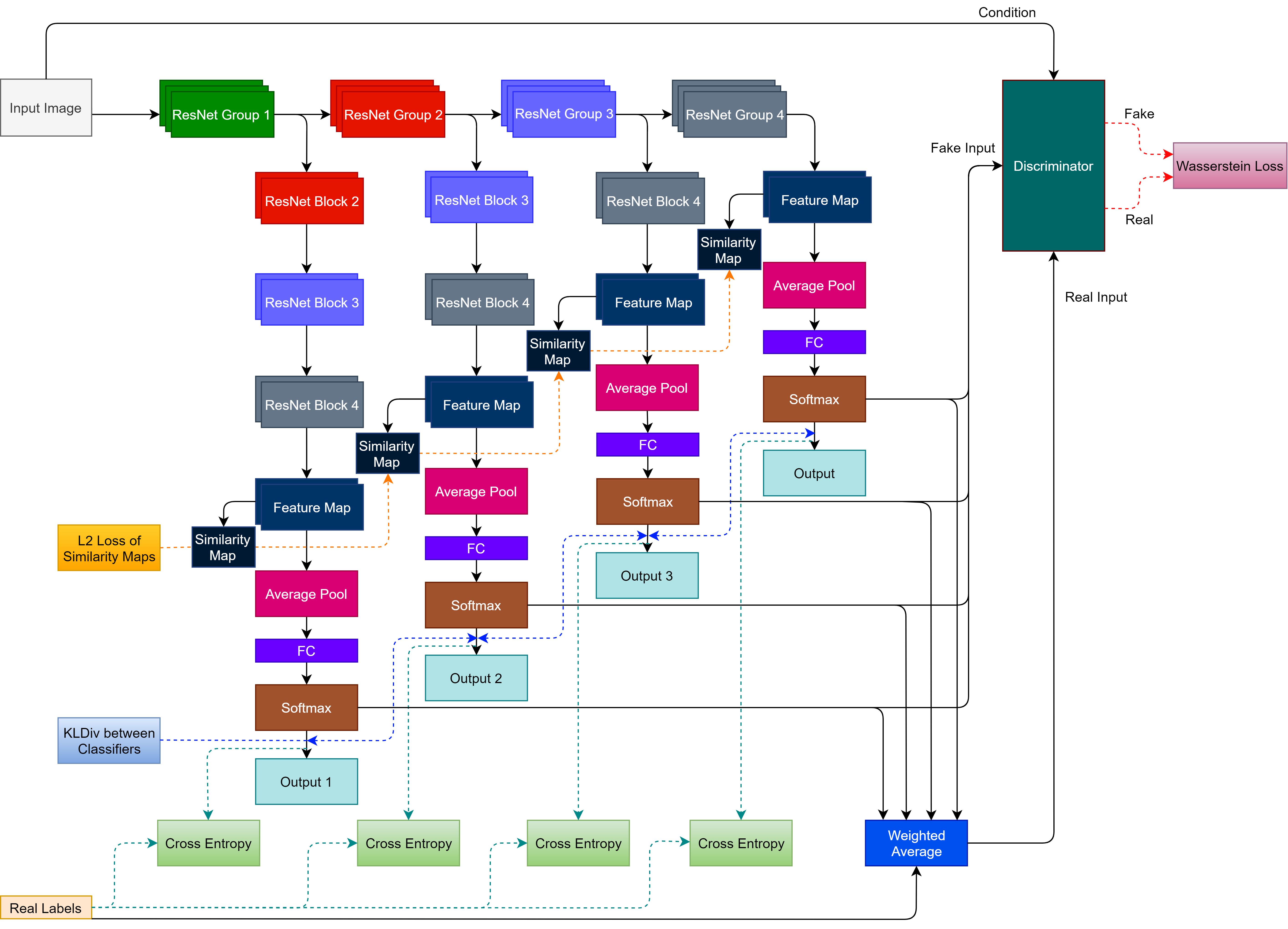}
		\label{fig:class_arch:a}
	}
	\hfil
	\subfloat[Discriminator architecture]{
		\includegraphics[width=1.6in]{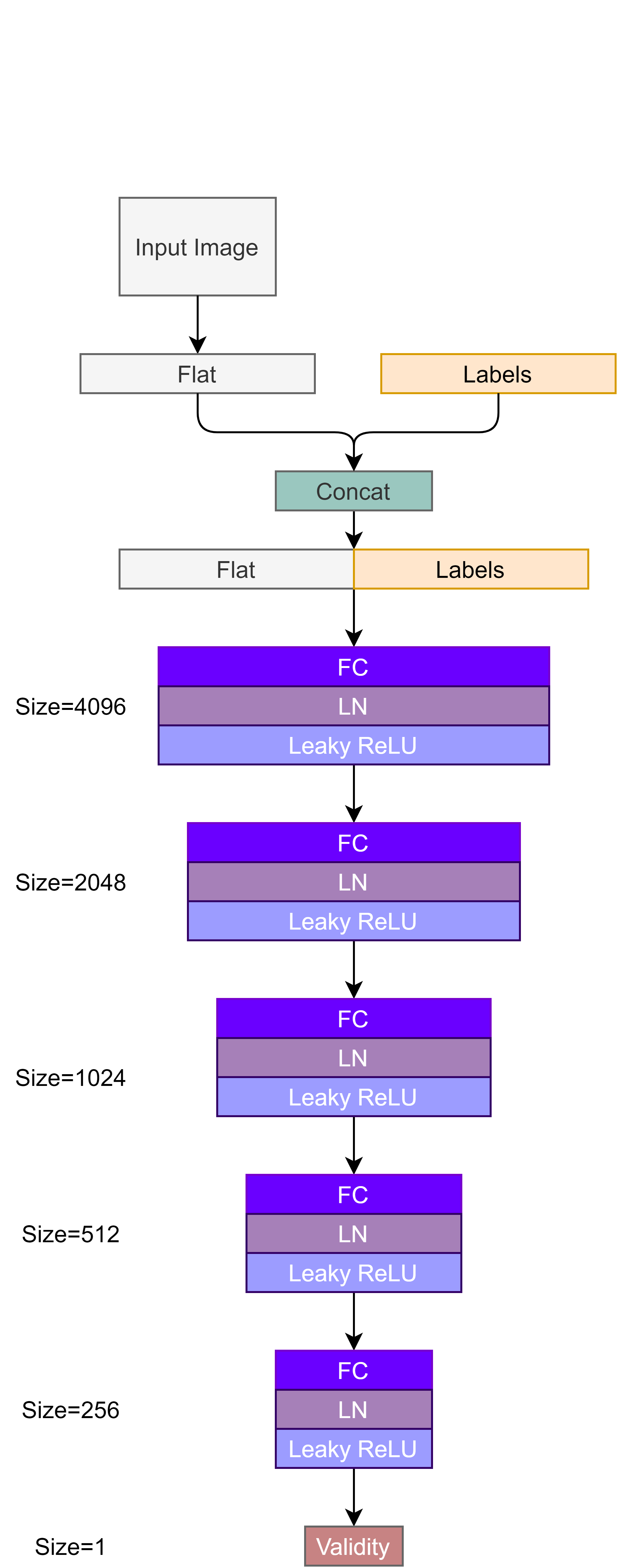}
		\label{fig:class_arch:b}
	}
	\vfill
	\subfloat{
    	\includegraphics[width=4.5in]{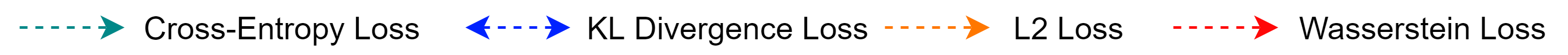}
    	\label{fig:class_arch:c}
	}
	\caption{(a) Proposed network architecture for image classification assembled on top of a ResNet model. Three additional classifiers are constructed after each ResNet group.  ResNet groups of each branch utilize the same ResNet blocks with a different number of convolutional layers. ResNet groups with identical colors use the same number of convolutional filters. For the final label prediction, each classifier is utilized, independently. (b) Proposed discriminator architecture. [FC: Fully Connected, LN: Layer Normalization]}
	\label{fig:class_arch}
\end{figure*}

\begin{figure*}[!t]
	\centering
	\subfloat[Proposed semantic segmentation architecture]{
		\includegraphics[width=5in]{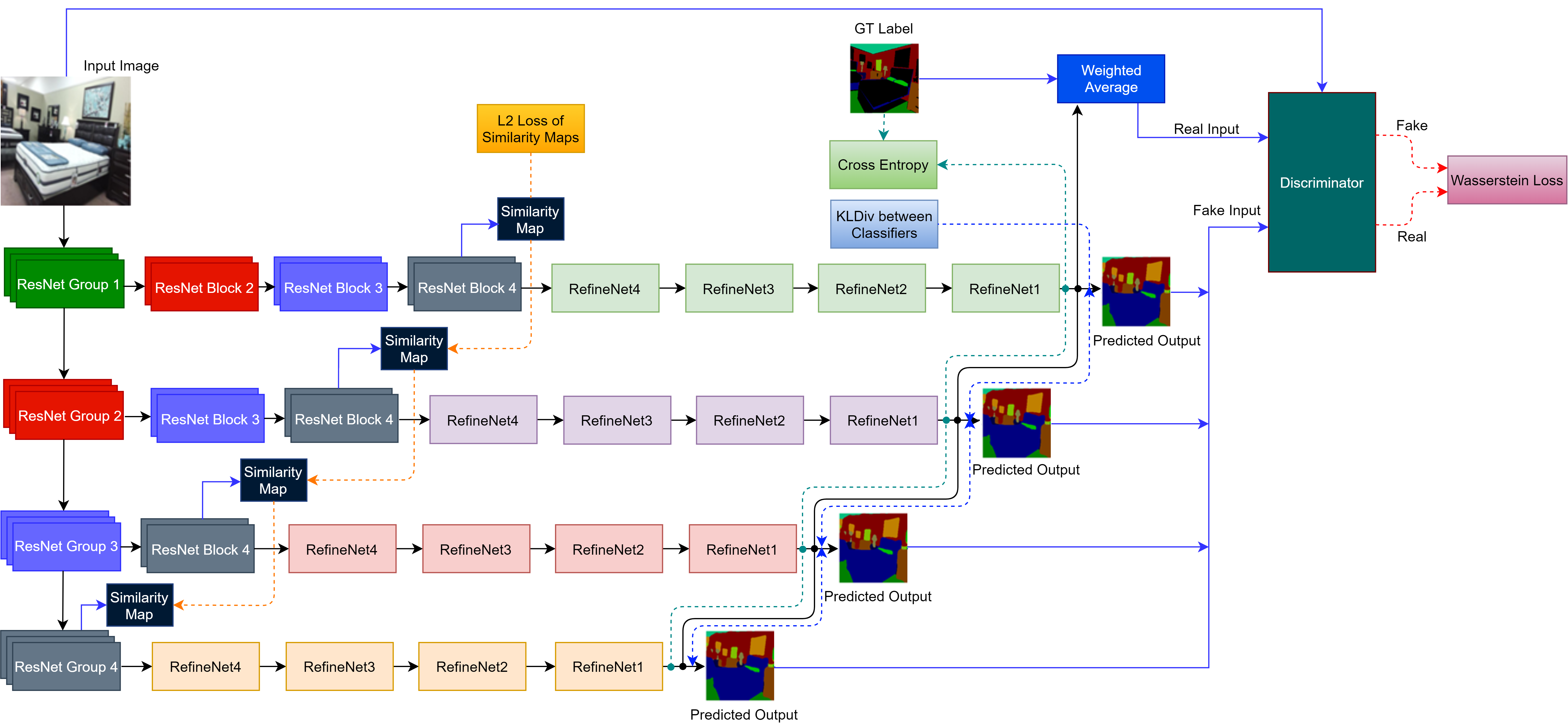}
		\label{fig:ss_arch:a}
	}
	\hfil
	\subfloat[Discriminator]{
		\includegraphics[width=1.0in]{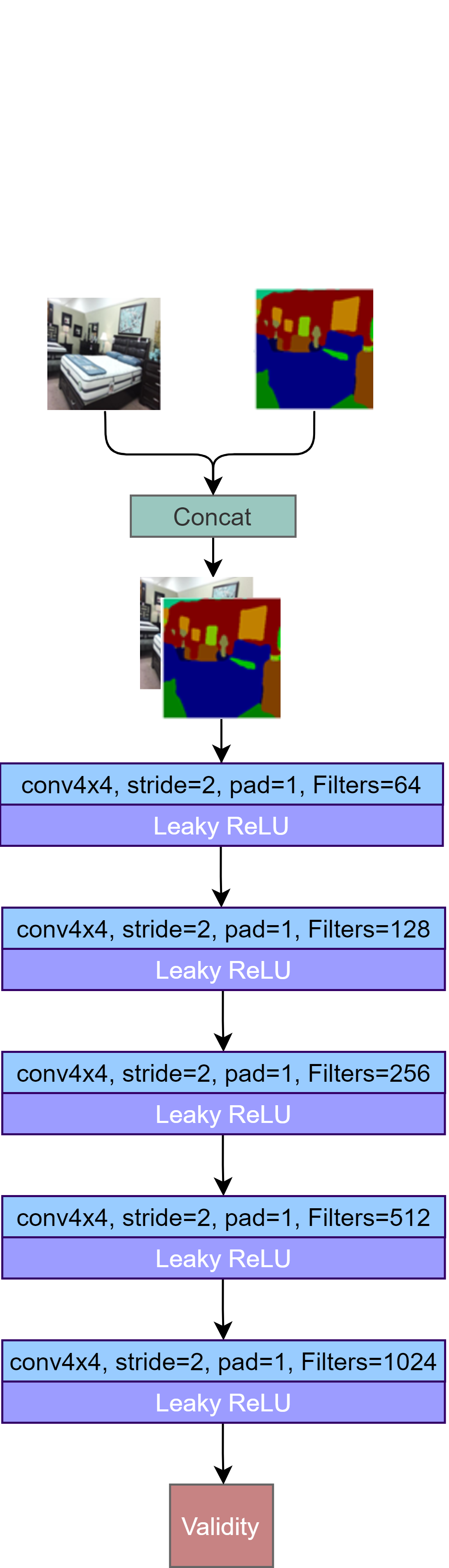}
		\label{fig:ss_arch:b}
	}
	\vfill
	\subfloat{
	\includegraphics[width=4.5in]{Img/Hint.png}
	\label{fig:ss_arch:c}
	}
	\caption{(a) Proposed network architecture for semantic segmentation assembled on top of a Light-Weight RefineNet model. Three additional segmenters are constructed after each ResNet group of the encoder. The same decoder is used for all branches. The similarity map is calculated for the last layer feature map of each encoder part. For the final label prediction, each sub-model can be utilized independently. (b) Proposed discriminator architecture.}
	\label{fig:ss_arch}
\end{figure*}

\section{Proposed Method}
\label{sec:3}
The proposed method has been focused on improving the performance of CNN models via ensembles of sub-neural network models. Two main intentions of the proposed architecture are i) to achieve the ensemble learning goals that the model universally performs better than each sub-model and ii) sub-models help each other to produce better results. In other words, ensemble learning aims at improving the model accuracy by aggregating predictions of multiple sub-models or classifiers. It is proposed that sub-models can assist each other by transferring and distilling their knowledge among themselves. Furthermore, these ideas can be applied in the adversarially learning framework to reinforce the amount of transferred knowledge by self-training or teacher-student training paradigms. The proposed model has three main steps: i) constructing and learning different sub-models, ii) transferring knowledge among each sub-model, and iii) utilizing adversarial learning to reinforce the KD. The overall architecture of the proposed models for image classification and semantic segmentation are illustrated in Fig. \ref{fig:class_arch}-a and Fig. \ref{fig:ss_arch}-a. In the following sub-sections, the design details are given.

\subsection{Sub-model constructing and branching}
The first step is making some changes in the primary network to construct some sub-models as classifiers. Based on the network architecture, several branches, $K$, are added to the model. Hence, these $K$ branches contain the same blocks as the primary network architecture, but the number of layers may differ (always less than the primary stream). These different classifiers partially share the same backbone where it can lead to better performance. 

The richest source for training the network is the labels' guidance.
Therefore, the principle loss function is the cross-entropy loss computed between the softmax layer's output of each classifier and real labels. The output of the softmax layer is computed as follows
\begin{equation}
\label{eq1}
p_k^i=\frac{\exp(a_k^i)}{\sum_{j}\exp(a_k^j)}
\end{equation}
where $p_k^i$ denotes the $i$-th class probability of the $k$-th classifier and $a_k^i$ denotes the logits of the $k$-th classifier.
The loss function is defined by summing over the losses of each classifier, as
\begin{equation}
\label{eq2}
\mathcal{L}_{CE} = \sum_{k=1}^{K+1}CrossEntropy(p_k, y)
\end{equation}
where $y$ denotes the true labels of images in the image classification task and the true semantic map in the semantic segmentation task. Also, $p_k$ denotes the probability vector of the softmax layer's output of the $k$-th classifier and $K$ denotes the number of attached branches.

\subsection{Transferring knowledge between sub-models}
In the proposed model, each classifier can promote other classifiers by transferring its knowledge. As such, there are other valuable supervision sources for training the whole network and improving its performance. Two resources are utilized for this purpose. The first one is the output of each classifier that represents a softer target for another classifier to learn its distilled knowledge. Therefore, minimizing a metric like the Kulback-Leibler (KL) divergence that measures the distance of two probability distributions can be considered between the output of each pair of classifiers to transfer the knowledge. 
The second one is the pair-wise similarities among feature maps of each sub-model that is utilized as another remarkable supervision for transferring knowledge among sub-models.  In the proposed method, the last convolutional layer of each sub-model of image classification architecture, as well as the last ones in the encoder part of the semantic segmentation network have been exploited.  
Consequently, computing similarity maps from these feature maps is supposed to obtain the importance of the relationships among different feature vectors. Therefore, the proposed supervision has been considered by making the shallower classifiers' similarity maps more similar to the deeper ones by only back-propagating the gradients in the direction of shallower classifiers. As such, minimizing a dissimilarity metric like the L2 distance can lead to this goal. In the following subsection, these two sources of knowledge transfer are explained mathematically.

\subsubsection{KL divergence loss}
To transfer the knowledge among each classifier, the output of softmax layers of each classifier is utilized, as they are softened targets and are easier to learn than actual labels. Now, each classifier can learn the distilled knowledge from all of the other softmax layer's output. The KL divergence loss is computed between these softmax layer's output of each pair of classifiers. This learning approach makes the output probability distribution of classifiers to be more similar to each other. The softened output of the softmax layer for knowledge distillation is computed by
\begin{equation}
\label{eq3}
q_k^{i,T} = \frac{\exp(a_k^i/T)}{\sum_{j}\exp(a_k^j/T)}
\end{equation}
where $T$ is the distillation temperature, $q_k^{i,T}$ denotes the softened $i$-th class probability of the $k$-th classifier with temperature $T$, and $a_k^i$ denotes the logits of the $k$-th classifier.
The loss function is defined by summing over the mean of KL divergence losses for each classifier, as
\begin{equation}
\label{eq4}
\mathcal{L}_{KL} = \frac{1}{K}    \sum_{i=1}^{K+1} \sum_{j = 1, j \neq i}^{K+1} KL(q_i^T || q_j^T)
\end{equation}
in which $q_i^T$ denotes the softened probability vector of the softmax layer's output of the $i$-th classifier and $KL(\cdot||\cdot)$ is the KL divergence distance of the vectors in image classification and semantic maps in the semantic segmentation model.

\subsubsection{L2 loss between similarity maps}
The proposed sub-model construction attaches branches to the primary model with a different number of layers.
Enforcing shallower ones to learn the embedded knowledge in feature maps of deeper classifiers makes them more precise. It happens by letting the gradients to only back-propagate in the path of the shallower classifier during the training phase. The $C \times H \times W$  input image is processed through the network and is transformed to $K+1$ different $C' \times H' \times W'$ feature maps that are located before the softmax layers in image classification network or at the end of the encoder in semantic segmentation network. Then, the similarity map is computed for each of them, which indicates the similarity between all pairs of feature vectors in a feature map. The dimension of the similarity map is $N \times N$, where $N = H' \times W'$. The similarity map for each feature map is calculated by 
\begin{equation}
\label{eq5}
S_k = [s_{ij}^k]_{N \times N}, s_{ij}^k = \frac{\langle f_i^k, f_j^k \rangle}{\norm{f_i^k}_2 \norm{f_j^k}_2}
\end{equation}
where $S_k$ is the similarity map of the $k$-th classifier, $f_i^k$ is the $i$-th feature vector of the $k$-th classifier, $\langle \cdot, \cdot \rangle$ is the inner product of the vectors, and $\norm{\cdot}_2$ is the L2 norm of the vector. The loss function is defined by summing over the mean of L2 losses for each classifier, as
\begin{equation}
\label{eq6}
\mathcal{L}_{L2} = \frac{1}{N^2}\sum_{i=1}^{K+1} \frac{1}{K+1-i} \sum_{j > i}^{K+1}\norm{S_i - S_j}_2^2.
\end{equation}

\subsection{Adversarial learning }

To exploit the generative adversarial learning, a conditional GAN architecture is utilized. For this purpose, a simple yet effective discriminator module is added to the network. Therefore, the whole training process converts to training an improved Wasserstein GAN (WGAN-GP) \cite{gulrajani2017improved} to enjoy this benefit.  Each classifier is considered as a generator; therefore, there are $K+1$ different generators with a simple discriminator to make the adversarial learning possible.  The discriminator consists of a simple sequence of fully connected layers following the layer normalization with the Leaky-ReLU activation function for the image classification task. For the semantic segmentation task, it contains several convolutional layers following Leaky-ReLU activations . Fig. \ref{fig:class_arch}-b and Fig. \ref{fig:ss_arch}-b show the architecture of the proposed discriminators, respectively. All of the generators get trained adversarially against the discriminator. The outputs of their softmax layers are considered as fake inputs for the discriminator. The weighted average over the ensemble of the classifiers and real labels are deemed to be the real input of the discriminator which is computed as 
\begin{equation}
\label{eq7}
r = \frac{\mu_r}{K+1} \sum_{i=1}^{K+1}p_i + (1 - \mu_r)y
\end{equation}
where $\mu_r$ is a hyper-parameter to set the effect of the real labels and ensemble, $p_i$ denotes the output of the $i$-th generator, and $y$ denotes the ground-truth. The corresponding input image is used as the input condition for the discriminator. The discriminator trains with WGAN-GP loss function \cite{gulrajani2017improved}, computed for $K+1$ generators as
\begin{equation}
\label{eq8}
\begin{split}
\mathcal{L}_D = \frac{1}{K+1}\sum_{i=1}^{K+1} \mathbb{E}_{p_i \sim \mathbb{P}_{g_i}}[D(p_i|I)] - \mathbb{E}_{r \sim \mathbb{P}_r}[D(r|I)] + \\*
\frac{\lambda_{GP}}{K+1} \sum_{i=1}^{K+1} \mathbb{E}_{\hat{p}_i \sim \mathbb{P}_{\hat{g}_i}}[(\norm{\nabla_{\hat{p}_i}D(\hat{p}_i|I)}_2 - 1)^2]
\end{split}
\end{equation}
where $I$ denotes the input image, $\mathbb{E[\cdot]}$ is the expectation operator, $D(\cdot|I)$ is discriminator's output given the input image, $\nabla$ is gradient operator, $\lambda_{GP}$ denotes a hyper-parameter to set the effect of the gradient penalty, $p_i \sim \mathbb{P}_{g_i}$ denotes the output of the $i$-th generator, $r \sim \mathbb{P}_{r}$ denotes the sample computed by the weighted average of the ensemble of all sub-models' output and ground-truth label, and $\hat{p}_i \sim \mathbb{P}_{\hat{g}_i}$ denotes a linear interpolation between the real label and the generator's output. The loss function of generators is defined in the same way of \cite{gulrajani2017improved}, by 
\begin{equation}
\label{eq9}
\mathcal{L}_{W} = -\frac{1}{K+1} \sum_{i=1}^{K+1} \mathbb{E}_{p_i \sim \mathbb{P}_{g_i}}[D(p_i|I)].
\end{equation}

\subsection{Overall model training}
To apply the superiority of all aforementioned ideas, the whole network is trained with  $\mathcal{L}_{SD}$ objective function,  defined as 
\begin{equation}
\label{eq10}
\mathcal{L}_{SD} = (1 -\alpha) \mathcal{L}_{CE} + \alpha \mathcal{L}_{KL} + \beta \mathcal{L}_{L2} + \gamma \mathcal{L}_{W}. 
\end{equation}

The hyperparameters $\alpha$, $\beta$, and $\gamma$ are chosen by searching manually in a specific range of values for each configuration and dataset. At the end of the training process, each classifier or segmenter can be utilized, independently. Therefore, the number of parameters and operations is less than or equal to the number of parameters and operations of the primary network base on the chosen classifier.

\begin{figure*}[!t]
	\centering
	\includegraphics[width=7in]{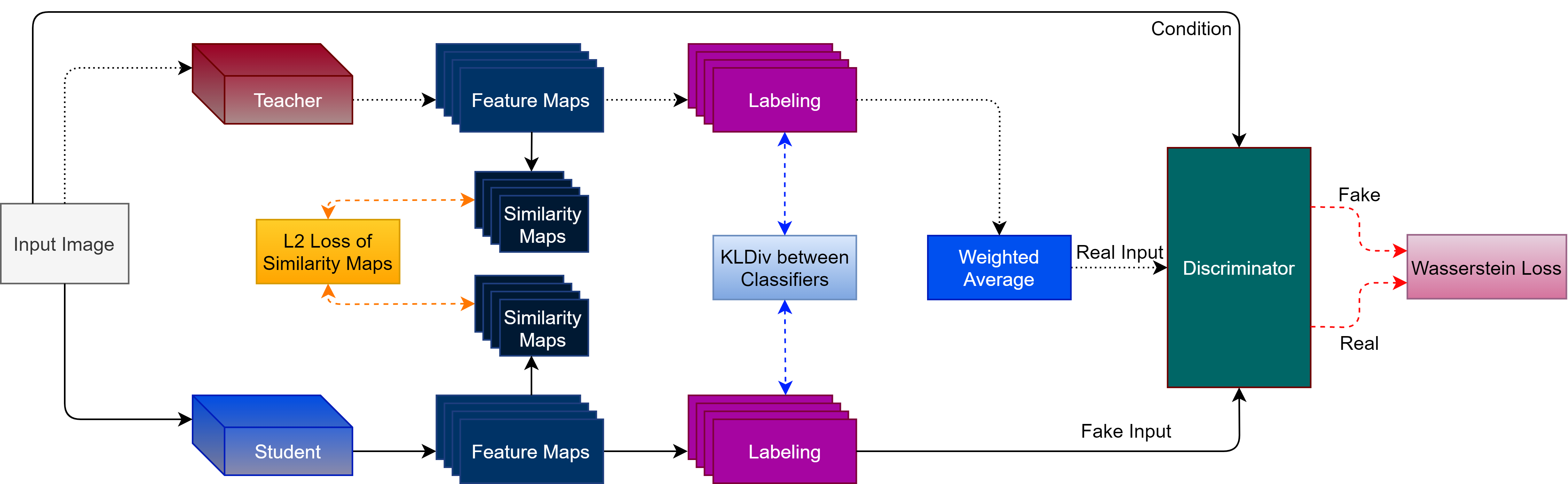}
	\caption{Proposed knowledge distillation network architecture. The pre-trained teacher model  has been utilized. 
	Each classifier in the student model learns from all teachers' classifiers.}
	\label{fig:kd_arch}
\end{figure*}

\section{Experimental Results}
\label{sec:4}
In this section, the effectiveness of the proposed method is evaluated on both image classification and semantic segmentation models. 

\subsection{Image classification architecture}
The proposed model has been evaluated on five main labeling benchmark datasets of MNIST \cite{lecun2010mnist}, Fashion-MNIST \cite{xiao2017fashion},  CIFAR-10 \cite{krizhevsky2009learning}, CIFAR-100 \cite{krizhevsky2009learning},  and Tiny ImageNet \cite{le2015tiny}.  To compare the performance of the proposed method, it is designed with well-known models (like ResNet \cite{he2016deep} and ResNeXt \cite{xie2017aggregated}).
 During the training phase, simple data augmentation and learning rate decay are employed. Stochastic Gradient Descent (SGD) optimizer is chosen for generator and discriminator. All of the optimizers have the same set of hyperparameters. Their learning rate starts from 0.1 and decreases to 0, in 200 epochs, by the cosine annealing algorithm \cite{loshchilov2016sgdr}. The weight decay and momentum are set to 0.0005 and 0.9, respectively. The mini-batch size is 128 or 64; based on the model configuration and dataset.  All of the experiments are performed on a single GPU. The internal network layers are adjusted based on the image size of each dataset. The model is implemented by the Pytorch library and it will be made publicly available on github\footnote{https://github.com/mtghorbani/AdvSD}. 

\subsubsection{Results on MNIST and Fashion-MNIST}
The MNIST and Fashion-MNIST datasets contain $28 \times 28$ grayscale images of handwritten digits and clothes categorized in 10 different classes, respectively.  There are 60k training and 10k testing images in each dataset.  Four different ResNet configurations are employed as the primary models of the proposed model. Based on the size of input images, three branches are constructed on these primary models. The results of various ResNet models on these two datasets are listed in Table \ref{tab:mnist}. All of the networks get trained by optimizing the loss function defined in Eq. \ref{eq10}. The best model of each classifier is chosen individually during the training process.

\begin{table}
	\centering
	\caption{Results of various ResNet models on MNIST and Fashion-MNIST(F-MNIST) datasets.}
	\label{tab:mnist}
	\scalebox{0.9}{
	\begin{tabular}{ccccc}
		\hline
		Model    & Dataset                  & Classifier 1 & Classifier 2 & Classifier 3 \\ \hline
		ResNet20 & \multirow{4}{*}{MNIST}   & 99.73        & 99.63        & 99.70        \\ 
		ResNet32 &                          & 99.67        & 99.69        & 99.69        \\ 
		ResNet44 &                          & 99.71        & 99.73        & 99.74        \\ 
		ResNet56 &                          & 99.67        & 99.67        & 99.67        \\ \hline
		ResNet20 & \multirow{4}{*}{F-MNIST} & 94.19        & 94.29        & 94.38        \\ 
		ResNet32 &                          & 94.36        & 94.43        & 94.40        \\ 
		ResNet44 &                          & 94.32        & 94.31        & 94.32        \\ 
		ResNet56 &                          & 94.39        & 94.42        & 94.20        \\ \hline
	\end{tabular}}
\end{table}

\begin{table}[!t]
	\centering
	\caption{Results of different ResNet models on CIFAR-10 dataset. Baseline accuracies are reported by \cite{he2016deep}.}
	\label{tab:cifar10}
	\scalebox{0.85}{
	\begin{tabular}{cccccc}
		\hline
		\multicolumn{1}{c}{Model}    & Baseline & Classifier 1 & Classifier 2 & Classifier 3 & Classifier 4 \\ \hline
		ResNet20                       & 91.25    & 92.61        & 92.86        & 93.00        & -            \\ 
		ResNet32                       & 92.49    & 94.27        & 94.36        & 94.25        & -            \\ 
		ResNet44                       & 92.83    & 94.48        & 94.58        & 94.68        & -            \\ 
		ResNet56                       & 93.03    & 94.67        & 94.77        & 94.83        & -            \\ 
		\multicolumn{1}{c}{ResNet110} & 93.57   & 94.73        & 94.94        & 95.14        & -            \\ 
		\multicolumn{1}{c}{ResNet34} & -        & 95.54        & 96.05        & 95.93        & 95.94        \\ 
		\multicolumn{1}{c}{ResNet50} & -        & 92.57        & 94.66        & 94.48        & 94.69        \\ \hline
	\end{tabular}}
\end{table}

\subsubsection{Results on CIFAR-10 and CIFAR-100}
CIFAR-10 and CIFAR-100 are datasets of $32 \times 32$ RGB images. Each of them contains 50k training and 10k testing images assigned to 10 and 100 classes, respectively. Simple data augmentations (like random crop, random horizontal flip, and normalization) are applied on training images. Tables \ref{tab:cifar10} and  \ref{tab:cifar100} show the validation result of various networks on CIFAR-10 and CIFAR-100, respectively. To illustrate the effectiveness of the proposed model, the results of the baseline model have been reported. The baseline model is the proposed model without branching and employing different proposed loss functions. The output of each branched classifier has improved compared with the baseline model.  For the first five models of these two tables, Classifier 3 is the primary network. For the rest of the models, Classifier 4 is the primary network. 
The gained improvement achieved by the proposed model varies between 1.56\% for ResNeXt29\_8x64d and 4.76\% for ResNet50 on CIFAR-100.
 
\begin{table}[!t]
	\centering
	\caption{Results of different models on CIFAR-100 dataset.}
	\label{tab:cifar100}
	\scalebox{0.8}{
	\begin{tabular}{cccccc}
		\hline
		Model              & Baseline & Classifier 1 & Classifier 2 & Classifier 3 & Classifier 4 \\ \hline
		ResNet20           & -        & 67.62        & 68.86        & 69.82        & -            \\ 
		ResNet32           & -        & 72.18        & 72.82        & 73.11        & -            \\ 
		ResNet44           & -        & 74.37        & 74.44        & 75.23        & -            \\ 
		ResNet56           & -        & 74.84        & 75.39        & 75.73        & -            \\ 
		ResNeXt29\_8x64d   & 82.23 \cite{xie2017aggregated}    & 83.18        & 83.20        & 83.79        & -            \\ 
		ResNet18           & 77.09  \cite{zhang2019your}  & 79.72        & 80.47        & 81.26        & 81.38        \\ 
		ResNet34           & -        & 81.30        & 81.66        & 81.81        & 81.79        \\ 
		ResNet50           & 77.68 \cite{zhang2019your}   & 81.32        & 81.86        & 82.02        & 82.44        \\ \hline
	\end{tabular}}
\end{table}

The results of different methods for self-distillation are listed in Table \ref{tab:comparesd}. The proposed method outperforms earlier ideas with a significant margin. There is a 3\% improvement than deeply supervised networks \cite{lee2015deeply}, 2.74\% than the  self-distillation method of \cite{zhang2019your}, and 1.12\% than multi-self-distillation technique \cite{luan2019msd} for ResNet18 primary model on CIFAR-100. The enhancement of the proposed method over the three mentioned models with the architecture of ResNet50  is 2.17\%, 1.88\%, and 0.66\%, respectively. Based on these results, converting the training process to an adversarial learning framework by utilizing a discriminator with the conditional WGAN-GP \cite{gulrajani2017improved} loss function alongside the cross-entropy loss, KL divergence loss between all pairs of sub-models, and L2 loss between similarity maps (instead of feature maps) can boost the accuracy up to 3\% improvement.

\begin{table}[!t]
	\centering
	\caption{Results of different self-distillation methods on CIFAR-100. [DSN: deeply supervised nets \cite{lee2015deeply}, SD: self-distillation of \cite{zhang2019your}, Multi\_SD:  multi-self-distillation technique \cite{luan2019msd}] }
		\label{tab:comparesd}
	\scalebox{0.8}{
	\begin{tabular}{ccccccc}
		\hline
		Model                     & Baseline               & Method        & Classifier 1   & Classifier 2   & Classifier 3   & Classifier 4   \\ \hline
		\multirow{4}{*}{ResNet18} & \multirow{4}{*}{77.09} & DSN           & 67.23          & 73.80          & 77.75          & 78.38          \\ 
		&                        & SD      & 67.85          & 74.57          & 78.23          & 78.64          \\ 
		&                        & Multi\_SD     & 78.93          & 79.63          & 80.13          & 80.26          \\ 
		&                        & \textbf{Ours} & \textbf{79.72} & \textbf{80.47} & \textbf{81.26} & \textbf{81.38} \\\hline
		\multirow{4}{*}{ResNet50} & \multirow{4}{*}{77.68} & DSN           & 67.87          & 73.80          & 74.54          & 80.27          \\ 
		&                        &SD      & 68.23          & 74.21          & 75.23          & 80.56          \\ 
		&                        & Multi\_SD     & 78.60          & 80.36          & 81.67          & 81.78          \\ 
		&                        & \textbf{Ours} & \textbf{81.32} & \textbf{81.86} & \textbf{82.02} & \textbf{82.44} \\ \hline
	\end{tabular}}
\end{table}

\subsubsection{Results on Tiny ImageNet}
The Tiny ImageNet contains 100k training, 10k validation, and 10k testing of $64 \times 64$ RGB images classified in 200 different classes. There are 500 training, 50 validation, and 50 testing images per class. In the training phase, several data augmentation operations (such as random rotation, translation, horizontal flip, contrast changing, and normalization) are applied on images. The results of  different configurations of ResNet model on the validation set are reported in Table \ref{tab:tinyimagenet}.  Classifier 4 and 3 are the primary network based on the structures of residual building blocks. It is worth  mentioning  that the accuracy of selected baseline models have not been reported by other works on the Tiny ImageNet dataset.

\begin{table}[!t]
	\centering
	\caption{Results of different ResNet models on Tiny ImageNet dataset.}
	\label{tab:tinyimagenet}
	\scalebox{0.9}{
	\begin{tabular}{cccccc}
		\hline
		Model            & Baseline & Classifier 1 & Classifier 2 & Classifier 3 & Classifier 4 \\ \hline
		ResNet32         & -        & 51.99        & 52.39        & 52.68        & -            \\ 
		ResNet44         & -        & 54.37        & 54.75        & 55.19        & -            \\ 
		ResNet18         & -        & 63.44        & 64.08        & 64.87        & 65.67        \\ 
		ResNet34         & -        & 67.39        & 67.53        & 67.73        & 68.23        \\ \hline
	\end{tabular}}
\end{table}


\subsection{Semantic segmentation architecture}
The Light-Weight RefineNet model (LW-RefineNet) \cite{nekrasov2018light} is selected as the primary model of the proposed method for the semantic segmentation task. Two different configurations of ResNet as the encoder of LW-RefineNet models are evaluated on the NYU-V2 \cite{Silberman:ECCV12} and CamVid \cite{brostow2009semantic} datasets. In the training phase, the learning rate decay and simple data augmentations (such as mirroring and the random cropping) are utilized. The SGD optimizer is chosen for the discriminator and both decoder and encoder parts of the generator. The networks are trained for 300 epochs on the NYU-V2 and 150 epochs on the CamVid dataset where the learning rates of optimizers decreases every 100 and 50 epoch, respectively. For the encoder, learning rates varies in [5e-4, 2.5e-4, 1e-4]. The learning rate of the decoder and discriminators are 10 and 100 times greater than the encoder, accordingly. The momentum and weight decay are set to 0.9 and 0.00001, respectively. A $500 \times 500$ crop of main images are utilized for both training and testing. The code is developed in the Pytorch library. All of the experiments are performed on a single GPU.

\subsubsection{Results on NYU-V2}
The NYU-V2 is a dataset of $480 \times 640$ RGB images and their densely labeled pairs of aligned RGB and depth images. It contains 795 training and 654 test images. Each image pixel  is categorized into one of 41 different classes (40 distinct objects). Table \ref{tab:nyu} lists the results of two different configurations of LW-RefineNet on this dataset. Baseline results of LW-RefineNet 18 and 34 are obtained by setting the ResNet-18 and ResNet-34 as encoders with the corresponding RefineNet blocks as their decoders. 
The obtained results show that significant improvements have been achieved via transferring knowledge among different segmenters that are branched over the original stream of the primary model.
Segmenter 4 is the primary model that is boosted by other segmenters trained adversarially in the proposed self-distillation paradigm.

\begin{table}[!t]
	\centering
	\caption{MIoU of different Light-Weight RefineNet models on NYU-V2 dataset.}
	\label{tab:nyu}
	\scalebox{0.8}{
	\begin{tabular}{cccccc}
	\hline
	Model   & \multicolumn{1}{l}{Baseline} & Segmenter 1 & Segmenter 2 & Segmenter 3   & \multicolumn{1}{l}{Segmenter 4} \\ \hline
	LW-RF18 & 36.7                          & 26.2        & 32.5        & 37.2 & 39.7                    \\ 
	LW-RF34 & 40.7                          & 30.0        & 34.6        & 41.1 & 41.9                    \\ \hline
	\end{tabular}}
\end{table}

\subsubsection{Results on CamVid}
The CamVid is a dataset of urban images consisting of $720 \times 960$ RGB images and their semantic map. It contains 367 training, 101 validation, and 233 test images. The labels are among 32 different categories (31 distinct objects). But, only 12 categories are mostly used (11 distinct objects). The results of two different configurations of LW-RefineNet on this dataset are depicted in Table \ref{tab:camvid}. Baseline network configurations are consisting of ResNet 18 and 34 as encoders and similar RefineNet blocks as decoders. Reported results are obtained by evaluating the best model of each branch on the validation set. The MIoU of test sets on the best model are reported. The proposed method improves the MIoU of the primary model by a remarkable margin. Even earlier branches, which are shallower and have fewer parameters and operations, obtain better results than the baseline model.

\begin{table}[!t]
	\centering
	\caption{MIoU of different Light-Weight RefineNet models on test images of CamVid dataset.}
	\label{tab:camvid}
	\scalebox{0.8}{
	\begin{tabular}{cccccc}
	\hline
	Model   & \multicolumn{1}{l}{Baseline} & Segmenter 1 & Segmenter 2 & Segmenter 3   & \multicolumn{1}{l}{Segmenter 4} \\ \hline
	LW-RF18 & 59.7                          & 57.7        & 59.4        & 58.9 & 60.3                    \\ 
	LW-RF34 & 59.9                          & 60.4        & 59.5 & 60.9 & 62.3                    \\ \hline
	\end{tabular}}
\end{table}

\subsection{Teacher-student knowledge distillation results}

The proposed method is also exploited in the typical teacher-student framework. To reach this goal, both teacher and student network architectures are converted to a multi-branched system, as depicted in Fig.\ref{fig:class_arch}-a. Then, the teacher model was trained by the proposed method. Afterward,  the proposed loss sources are considered between the student model and teacher network. In other words, the KL divergence loss between the output of each pair of sub-models of student and teacher, L2 loss between similarity maps of shallower branches of student and deeper ones of teacher, and adversarial loss between student and teacher predictions are employed. As such, the network is designed by both proposed SD method and KD techniques. The final goal is minimizing an objective function $\mathcal{L} =  \mathcal{L}_{KD} + \mathcal{L}_{SD}$, where  $ \mathcal{L}_{KD}$, defined as
 \begin{equation}
\label{eq16}
\mathcal{L}_{KD} = \lambda_1 \mathcal{L}^{KD}_{KL} + \lambda_2 \mathcal{L}^{KD}_{L2} + \lambda_3 \mathcal{L}^{KD}_{W}
 \end{equation}
where $ \mathcal{L}^{KD}_{KL}$, $ \mathcal{L}^{KD}_{L2}$, and $ \mathcal{L}^{KD}_{W}$ are defined  among the branches of teacher and student model. $\lambda_i$ are the hyperparameters chosen by searching manually in a specific range of values for each configuration.
The overall architecture is shown in Fig. \ref{fig:kd_arch}.

To show the effect of the common teacher-student knowledge distillation framework, different ResNet networks were trained under the supervision of ResNet56 to achieve higher accuracy on the CIFAR-100 dataset. The results are presented in Table \ref{tab:kdsd}. There is a 1.89\% improvement for ResNet20 supervised by a pre-trained ResNet56 network rather than individually using the proposed method.   The results of Table \ref{tab:kdsd} show that  0.31\% improvement is achieved by utilizing the ResNet56  as both teacher and student model. Exploiting the same teacher network as the student model was firstly proposed in \cite{furlanello2018born}.

\begin{table}[!t]
	\centering
	\caption{Training various ResNet architectures using both proposed SD method and KD, under supervision of pre-trained ResNet56, on CIFAR-100 dataset. }
	\label{tab:kdsd}
	\begin{tabular}{ccccc}
    \hline
    Teacher                   & Student    & Classifier 1 & Classifier 2 & Classifier 3 \\ \hline
    \multirow{3}{*}{ResNet56} & ResNet20 & 68.77        & 69.73        & 71.71        \\ 
                              & ResNet32 & 72.81        & 73.51        & 73.47        \\
                              & ResNet56 & 75.01        & 75.94        & 76.04        \\ \hline
    \end{tabular}
\end{table}

\subsection{Ablation study: impact of different losses}
\begin{figure*}[!t]
	\centering
	\subfloat[RGB]{
		\includegraphics[width=.8in]{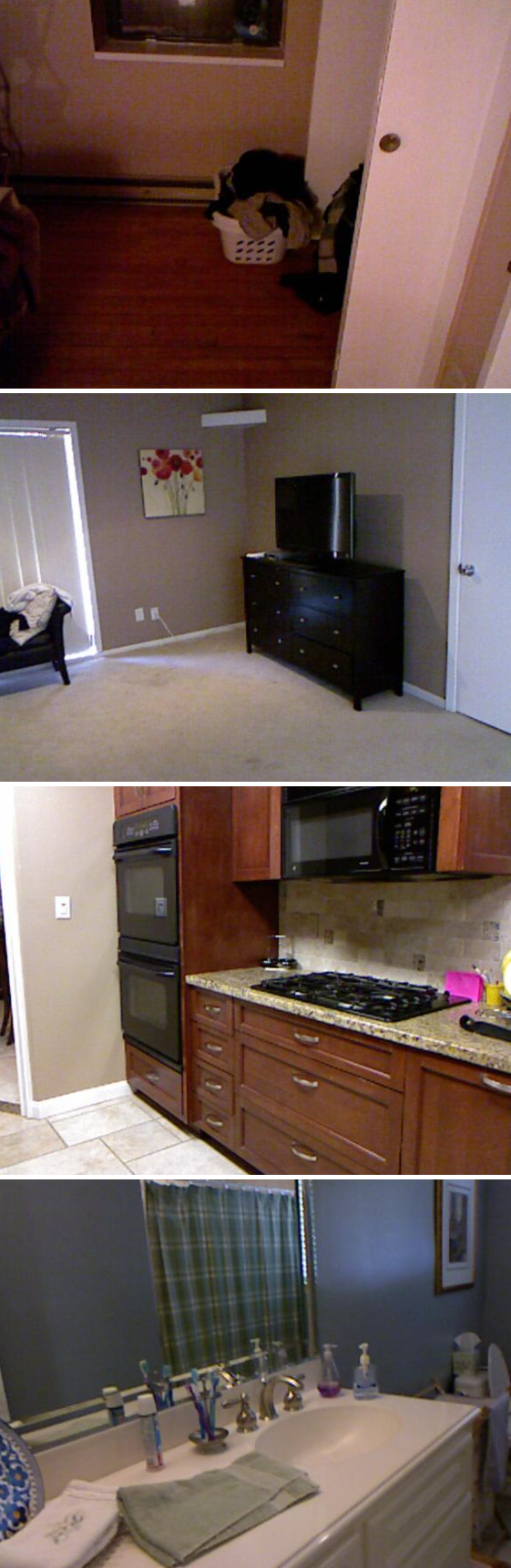}\hspace{0.001cm}
		\label{fig:losses_rgb}
	}
    \subfloat[GT]{
		\includegraphics[width=.8in]{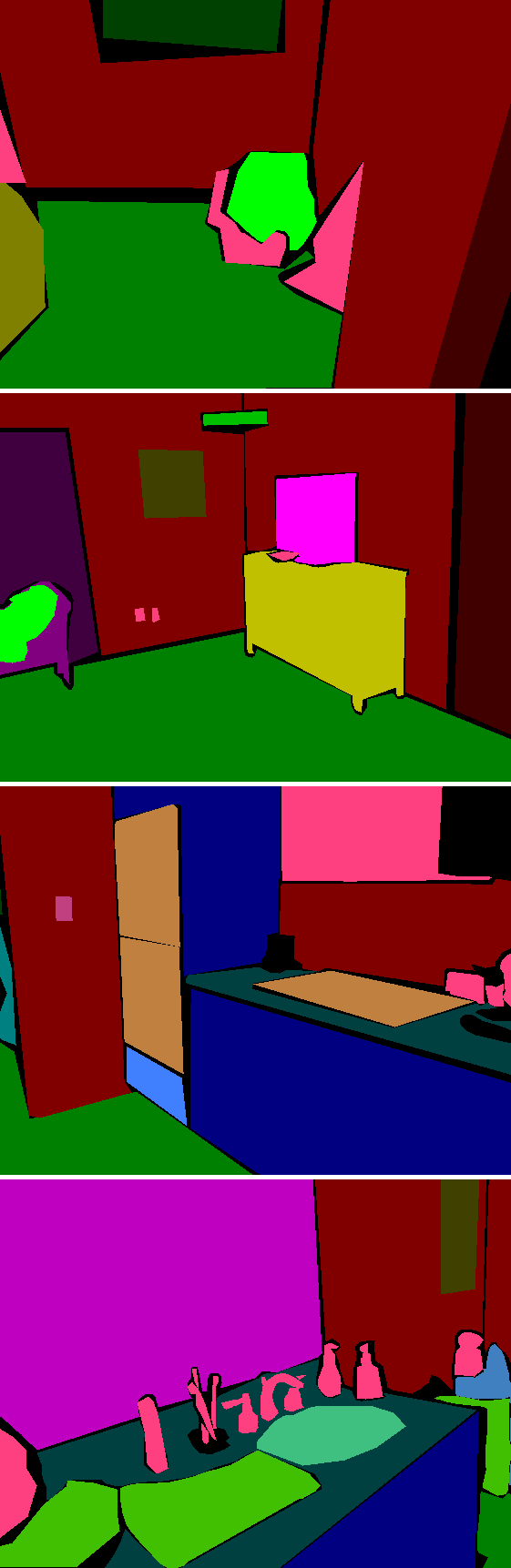}\hspace{0.001cm}
		\label{fig:losses_gt}
	}
	\subfloat[Baseline]{
		\includegraphics[width=.8in]{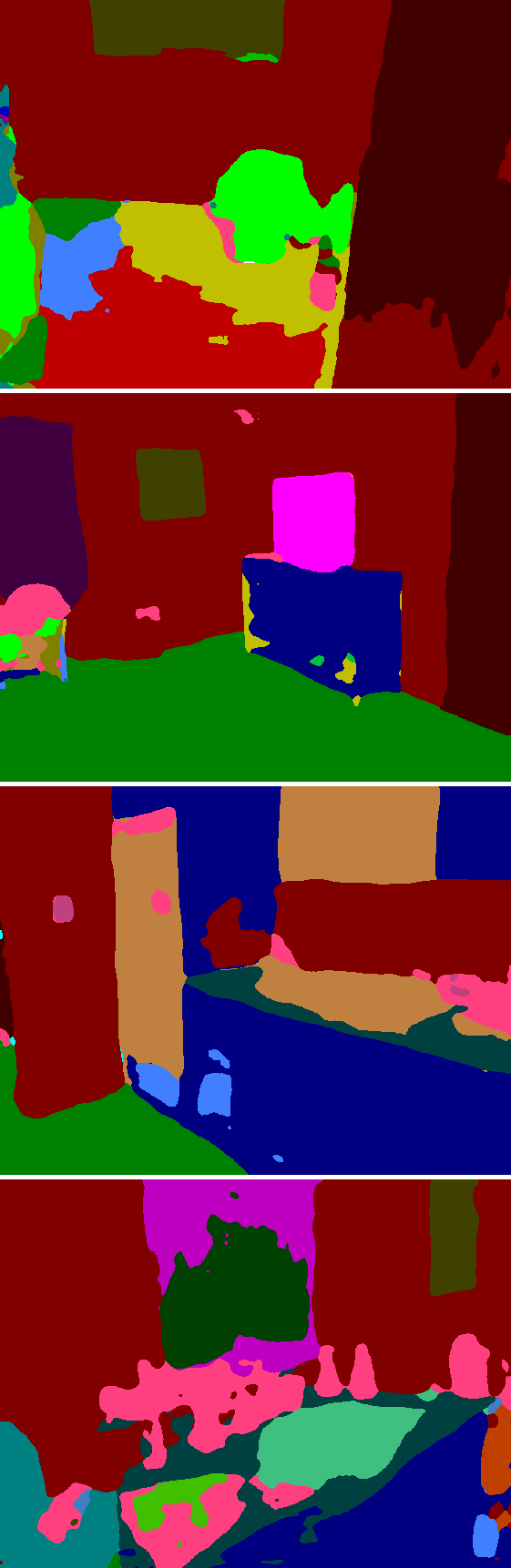}\hspace{0.001cm}
		\label{fig:losses_s}
	}
    \subfloat[CE]{
		\includegraphics[width=.8in]{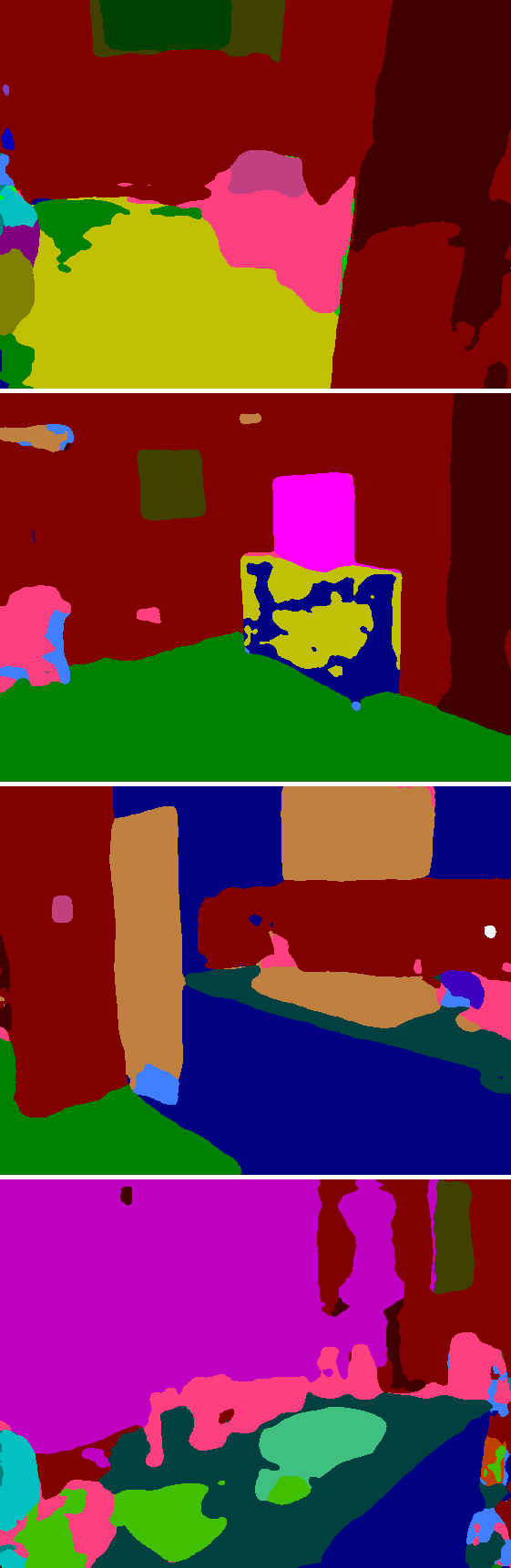}\hspace{0.001cm}
		\label{fig:losses_ce}
	}
    \subfloat[CE+KL]{
		\includegraphics[width=.8in]{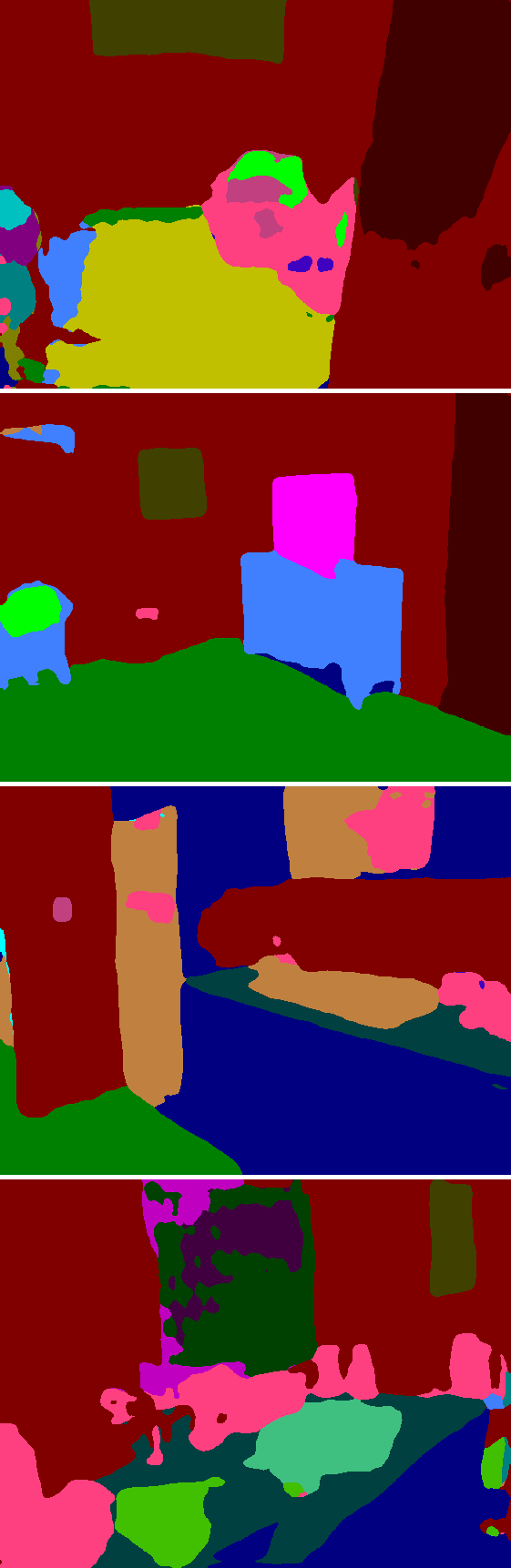}\hspace{0.001cm}
		\label{fig:losses_kl}
	}
    \subfloat[CE+KL+L2]{
		\includegraphics[width=.8in]{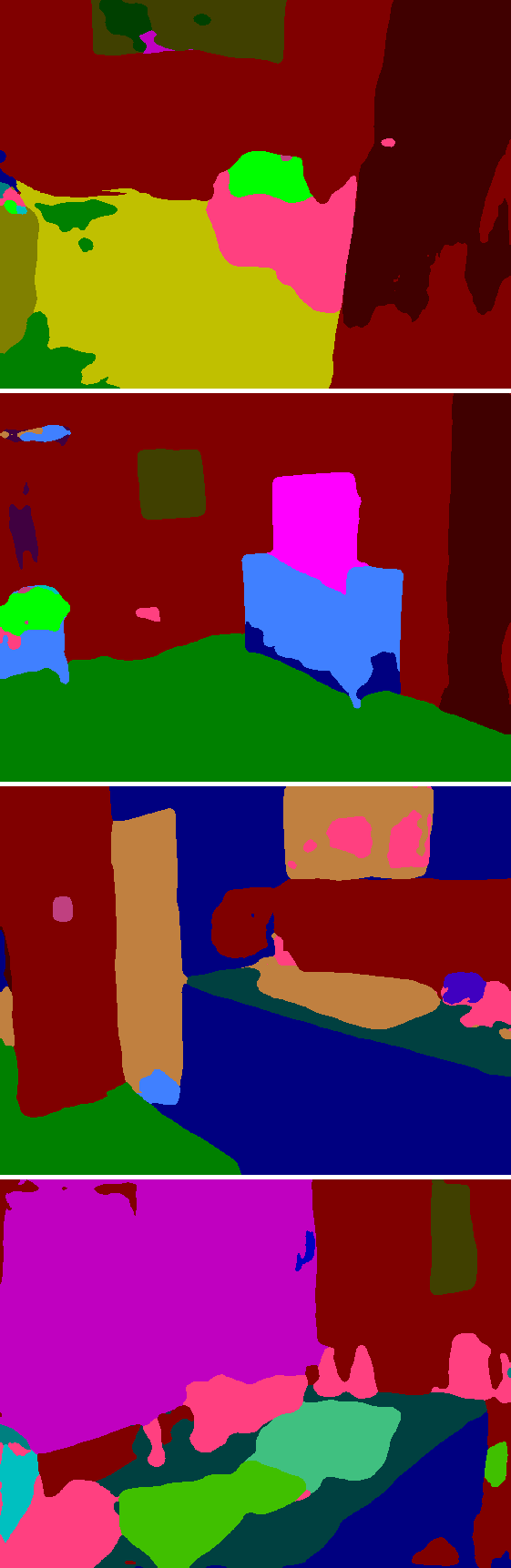}\hspace{0.001cm}
		\label{fig:losses_l2}
	}
    \subfloat[CE+KL+L2+W]{
		\includegraphics[width=.8in]{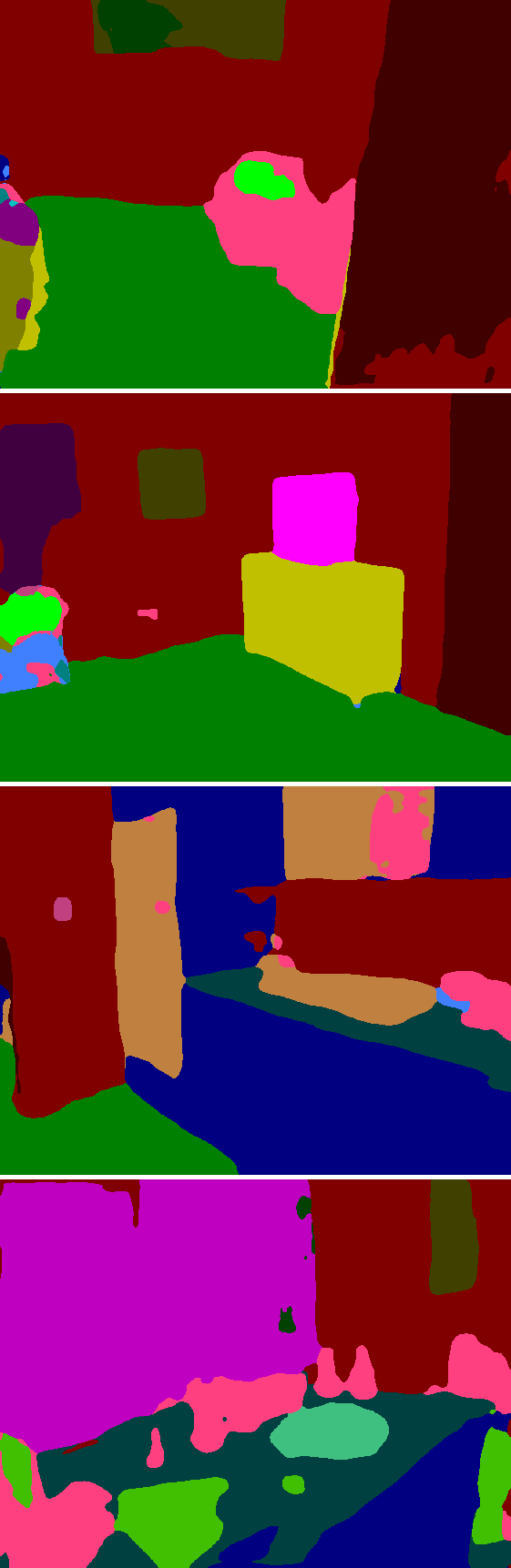}\hspace{0.001cm}
		\label{fig:losses_w}
	}
	\vfill
	\subfloat{
		\includegraphics[width=6.5in, height=.8in]{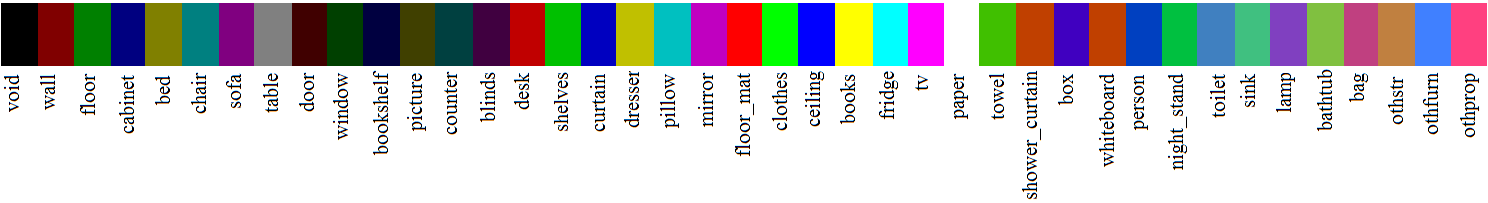}
		\label{fig:losses_colorbar}  
	}
	\caption{Effect of  each loss on Light-Weight RefineNet18 output.}
	\label{fig:losses}
\end{figure*}

Table \ref{tab:losses} shows the impact of adding losses step by step to different models on the CIFAR-100 and NYU-V2 dataset. Adding each loss function enhances the accuracy as it is expected. Starting with pre-trained ImageNet weights for the primary network also makes a slight improvement on image classification. Fig. \ref{fig:losses} shows the qualitative assessments of adding different loss sources on the output of LW-RefineNet18 on the NYU-V2 dataset. 

One of the remarkable properties of the proposed method is to improve the learning capacity of the lightweight models without increasing the number of computations at the inference time. Therefore, the proposed model is also compared with the baseline model in terms of accuracy, the number of parameters, and the Floating-Point Operations (FLOPs). Table \ref{tab:params} shows the number of parameters and FLOPS of a variety of models on CIFAR-100 and NYU-V2. The proposed method achieves higher accuracy with the same number of parameters and operations as the baseline model; at the cost of the computational power in the training phase. 

\begin{table*}[!t]
	\centering
	\caption{Impact of different loss functions. [KL: KL divergence loss between outputs, L2: L2 loss between similarity maps, W: Wasserstein loss in the GAN, and Pre: using pre-trained weights of ImageNet for the primary network]}
	\label{tab:losses}
	\begin{tabular}{l|c|c|cccc}
	\hline
	\multicolumn{1}{c}{Method} & \multicolumn{1}{l}{Dateset} & Baseline               & Classfier 1 & Classifier 2 & Classifier 3 & Classifier 4 \\ \hline
	ResNet18 + KL                & \multirow{4}{*}{CIFAR-100}   & \multirow{4}{*}{77.09} & 79.04       & 79.88        & 80.46        & 80.52        \\ 
	ResNet18 + KL + L2           &                              &                        & 79.15       & 79.79        & 80.70        & 80.88        \\ 
	ResNet18 + KL + L2 + W       &                              &                        & 79.16       & 79.67        & 80.90        & 81.15        \\ 
	ResNet18 + KL + L2 + W + Pre &                              &                        & 79.72       & 80.47        & 81.26        & 81.38        \\ \hline
	LW-RefineNet18               & \multirow{4}{*}{NYU}         & \multirow{4}{*}{36.7}  & 22.6        & 30.2         & 35.5         & 38.0         \\ 
	LW-RefineNet18 + KL          &                              &                        & 26.4        & 32.3         & 36.9         & 38.5         \\ 
	LW-RefineNet18 + KL + L2     &                              &                        & 25.8        & 32.4         & 37.7         & 39.2         \\ 
	LW-RefineNet18 + KL + L2 + W &                              &                        & 26.2        & 32.5         & 37.2         & 39.7         \\ \hline 
	LW-RefineNet34                 & \multirow{4}{*}{NYU}                     & \multirow{4}{*}{40.7}     & 24.8        & 33.6         & 40.3         & 41.1         \\ 
	LW-RefineNet34 + KL          &                              &                        & 28.7        & 35.0         & 40.9         & 41.5         \\ 
	LW-RefineNet34 + KL + L2     &                              &                        & 28.0        & 34.9         & 40.6         & 41.6         \\ 
	LW-RefineNet34 + KL + L2 + W &                              &                        & 30.0        & 34.6         & 41.1         & 41.9         \\ \hline
	\end{tabular}

\end{table*}

\begin{table*}[!t]
    \centering
    \caption{Performance evaluation of different models on CIFAR-100 and NYU-V2. }
    \label{tab:params}
    \begin{tabular}{ccccccc}
    \hline
    Model                   & Dataset                    & Accuracy & \# Train Params & \# Test Params & \# Train FLOPS & \# Test FLOPs \\ \hline
    ResNet18                & \multirow{6}{*}{CIFAR-100} & 77.09    & 11.2M           & 11.2M          & 556M           & 556M          \\
    ResNet18 (Ours)         &                            & 81.38    & 48.6M           & 11.2M          & 933M           & 556M          \\
    ResNet50                &                            & 77.68    & 23.7M           & 23.7M          & 1.301G         & 1.301G        \\
    ResNet50 (Ours)         &                            & 82.44    & 92.9M           & 23.7M          & 2.985G         & 1.301G        \\
    ResNeXt29\_8x64d        &                            & 82.23    & 34.5M           & 34.5M          & 5.396G         & 5.396G        \\
    ResNeXt29\_8x64d (Ours) &                            & 83.79    & 97.9M           & 34.5M          & 9.350G         & 5.396G        \\ \hline
    LW-RefineNet18          & \multirow{4}{*}{NYU-V2}    & 36.7     & 13.9M           & 13.9M          & 18.661G        & 18.661G       \\ 
    LW-RefineNet18 (Ours)   &                            & 39.7     & 38.1M           & 13.9M          & 61.321G        & 18.661G       \\
    LW-RefineNet34          &                            & 40.7     & 24.0M           & 24.0M          & 28.200G        & 28.200G       \\ 
    LW-RefineNet34 (Ours)   &                            & 41.9     & 70.1M           & 24.0M          & 84.079G        & 28.200G       \\ \hline
    \end{tabular}
\end{table*}

\section{Conclusion}
\label{sec:5}
In this work, a novel method based on the adversarial knowledge transfer in a multi-branched network was proposed for both image classification and semantic segmentation purposes.  To design a multi-branched model, the ensemble of sub-neural network models was constructed on the original stream of the primary model. The knowledge distillation strategies were applied within the ensemble model based on the four different loss functions. The effectiveness of the proposed method was shown by evaluating a variety of network configurations on well-known benchmarks for both image classification and semantic segmentation tasks. The proposed method achieved higher performance in terms of accuracy with the same number of parameters and operations as the baseline model. The experimental results showed that at least 1.56\% and 0.6\% improvements were achieved for image classification and semantic segmentation,  and up to 4.76\% and 3.0\% enhancement also can be obtained, respectively. The experimental results also illustrated up to 2.74\% improvement over the earlier self-distillation methods reported on image classification tasks.


%





\ifCLASSOPTIONcaptionsoff
  \newpage
\fi



%
\bibliographystyle{IEEEtran}
\bibliography{SelfKDAL}

%








\end{document}